\documentclass[lettersize,journal]{IEEEtran}
\usepackage{amsmath,amsfonts}
\usepackage{array}
\usepackage[caption=false,font=footnotesize]{subfig}

\usepackage{textcomp}
\usepackage{stfloats}
\usepackage{url}
\usepackage{verbatim}
\usepackage{graphicx}
\usepackage{booktabs}
\usepackage{cite}
\usepackage{amsmath}
\usepackage{paralist}

\usepackage{algorithmic}

\usepackage{color}
\usepackage{url}
\definecolor{indigo}{rgb}{0.0, 0.25, 0.42}
\definecolor{darkorange}{rgb}{1.0, 0.55, 0.0}
\definecolor{darkblue}{rgb}{0.122, 0.435, 0.698}

\usepackage{microtype}
\usepackage{graphicx}
\usepackage{booktabs} 

\usepackage[lined,boxed,commentsnumbered,ruled]{algorithm2e}
\usepackage{algorithmic}

\usepackage{amsmath}
\usepackage{amssymb}
\usepackage{mathtools}
\usepackage{amsthm}
\usepackage{stfloats}

\usepackage{amsmath}
\usepackage{amssymb}
\usepackage{mathtools}
\usepackage{amsthm}
\usepackage{stfloats}
\usepackage{hyperref}
\usepackage[capitalize,noabbrev]{cleveref}

\begin{document}
\title{JPEG Compliant Compression for Both Human and Machine, A Report}

\author{Linfeng~Ye~\IEEEmembership{Member,~IEEE,}
\IEEEcompsocitemizethanks{\IEEEcompsocthanksitem  L. Ye, is with the Department of Electrical and Computer Engineering, University of Waterloo, Waterloo, ON N2L 3G1, Canada (e-mail: l44ye@uwaterloo.ca).
}
}

\maketitle

\begin{abstract}
Deep Neural Networks (DNNs) have become an integral part of our daily lives, especially in vision-related applications. However, the conventional lossy image compression algorithms are primarily  designed for the Human Vision System (HVS), which can non-trivially compromise the DNNs' validation accuracy after compression, as noted in \cite{liu2018deepn}. Thus developing an image compression algorithm for both human and machine (DNNs) is on the horizon.

To address the challenge mentioned above, in this paper, we first formulate the image compression as a multi-objective optimization problem which take both human and machine prespectives into account, then we solve it by linear combination, and proposed a novel distortion measure for both human and machine,  dubbed Human and Machine-Oriented Error (HMOE). After that, we develop Human And Machine Oriented Soft Decision Quantization (HMOSDQ) based on HMOE, a lossy image compression algorithm for both human and machine (DNNs), and fully complied with JPEG format. In order to evaluate the performance of HMOSDQ, finally we conduct the experiments for two pre-trained well-known DNN-based image classifiers named Alexnet \cite{Alexnet} and VGG-16 \cite{simonyan2014VGG} on two subsets of the ImageNet \cite{deng2009imagenet} validation set: one subset included images with shorter side in the range of 496 to 512, while the other included images with shorter side in the range of 376 to 384. Our results demonstrate that HMOSDQ outperforms the default JPEG algorithm in terms of rate-accuracy and rate-distortion performance. For the Alexnet comparing with the default JPEG algorithm, HMOSDQ can improve the validation accuracy by more than $0.81\%$ at $0.61$ BPP, or equivalently reduce the compression rate of default JPEG by $9.6\times$ while maintaining the same validation accuracy.
\end{abstract}

\section{Introduction}\label{sec:intro}
\IEEEPARstart{I}{n} the present era, Deep Neural Networks (DNNs) have been deeply entrenched in our lives, and have became one of the indispensable parts in the computer vision-based algorithms \cite{ota2017deep,HAYAT2018198,10619241, SurveyDLforMultimedia,DLMultimediaReview,10900607,20240386275}. As one of the most fundamental computer vision problems, image classification has been used in diverse real-world applications involving both humans and machines, such as Google Lens and the diagnostic imaging \cite{ComputerVisionMedical}.

To alleviate the data storage and transfer burden, image compression algorithms are applied to reduce the file size. Therefore, the rate by which the data is compressed is of high importance. To measure the compression rate, Bits Per Pixel (BPP) is commonly used. As the \textit{de facto} human-oriented lossy image compression algorithms, JPEG and MPEG can efficiently reduce the redundancy of images, by first quantizing the Discrete Cosine Transform (DCT) coefficients of the raw images to DCT indices. Then, the resulting DCT indices are compressed using a lossless compression algorithm to further reduce the file size. However, the distortion introduced in quantization step can significantly degrade the accuracy of DNNs \cite{bhowmik2022lost, li2019adacompress,yang2025coded, 10.1007/978-3-031-73024-5_10}. For example, when the validation images are compressed by JPEG at 0.6 BPP, the VGG-16's top-1 validation accuracy will drop more than $3.5\%$ on the ImageNet-1K \cite{deng2009imagenet}, which demonstrates the challenge posed by lossy compression algorithms for DNN-based applications, and the urgency for more advanced compression algorithms that take DNNs' performance into account. 

In this paper, we present a new approach to establish a trade-off among compression rate, DNN accuracy, and human distortion. To this end, we formulate a multi-objective optimization problem to balance these metrics, and propose a new distortion measure to evaluate the distortion for humans and DNNs (machines). Then, we develop a lossy image compression algorithm that jointly optimizes run-length coding, Huffman coding, and quantization table to achieve a better trade-off among the three objectives mentioned above. The main contributions of this paper are summarized as follows:

\begin{itemize}    
\item We present a formulation of a multi-objective optimization problem that considers the challenge of establishing trade-off among DNN accuracy, compression rate, and human distortion. To factor in DNN accuracy, we introduced sensitive $S$, which reveals the relation between the distortion applied to the DCT coefficients of natural images, and the resulting change in surrogate loss value.

\item To estimate sensitivity offline for images with various resolutions, we propose Adaptive Sensitivity Mapping (ASM), a straightforward yet accurate linear mapping technique for approximating the estimated sensitivity $\hat{S}$ of images at one resolution using the $\hat{S}'$ of images at a lower resolution, based on moderate statistical assumptions.

\item With the help of $\hat{S}$ and ASM, we introduce a novel distortion measure dubbed Human and Machine Oriented Distortion (HMOD) to evaluate the distortion for humans and DNNs (machines).

\item Base on the HMOE, we then develop Human and Machine Oriented Soft Decision Quantization (HMOSDQ) algorithm. Which jointly optimized the run-length coding, Huffman coding and quantization table with JPEG format compatibility for both human and DNN model.

\item Experimental results presented in the paper show that HMOSDQ can achieve a better trade-off among DNN accuracy, human distortion, and compression rate. For Alexnet \cite{krizhevsky2017imagenet}, comparing with the default JPEG algorithm, HMOSDQ can improve the validation accuracy by more than $0.81\%$ at 0.61 BPP, or equivalently reduce the compression rate of default JPEG by $9.6\times$ while maintaining the same validation accuracy.
\end{itemize}
The rest of the paper is organized as follows. Section \ref{sec:RelatedWorks} reviews existing works which attempt to reduce the file size of the images while mitigate the negative impact on image quality or DNN accuracy. Section \ref{sec:preliminaries} briefly describes of the JPEG mechanism and the conventional rate-distortion trade-off problem for JPEG compression. Section \ref{sec:ProbosedMethod} discusses the details of the HMOSDQ algorithm we proposed in the paper. In Section \ref{sec:experiments} we conduct several experiments to evaluate rate-distortion-accuracy performance of HMOSDQ. Finally, conclusions are drawn in Section \ref{sec:ConclusionAndFutureWorks} and we discuss some future works.

\section{Related works} 

\label{sec:RelatedWorks}
The need for efficient transmission and storage of digital images have driven the continuous evolution of digital image compression techniques. In particular, extensive research has been conducted on JPEG-compliant image compression techniques that leverage the Human Vision System  (HVS) characteristics over the past three decays \cite{319829, SALAMAH2025110966, 150939, 821739, 6898872} due to its widespread popularity. 

After the rise of DNN-based vision applications following the revolutionary success of Alexnet \cite{krizhevsky2017imagenet} in ILSVRC-2012 \cite{Imagenet} competition, the primary consumers of the images have become DNN-based vision algorithms rather than human beings. However, conventional image compression algorithms designed for HVS can significantly degrade DNN-based models' performance which was first reported in \cite{liu2018deepn}. To address this issue, authors want to minimize the distortion apply on the coefficients that are most sensitive to DNN-based algorithms. However, the heuristic piece-wise linear quantization table named DeepN-JPEG they proposed in the paper lacks theoretic support and can not mitigate model performance degradation remarkably. After that, several works have attempted to develop new compression algorithms to balance accuracy and compression rate. In \cite{xie2019source, li2020optimizing} researchers propose novel quantization table optimization algorithms, which exploit the perception model of a DNN to compress its input without compromising their performance. Another branch of work leverages the capacity of DNN-based model, quantization steps and hardware-platform to develop a joint optimization algorithms \cite{hu2020starfish, ye2025towards}.  These approaches have been criticized as non-universal, requiring iterative retraining of both the model and quantization table for every new task. Despite this limitation, they can achieve state-of-the-art performance in the accuracy compression-rate trade-off. Finally, Selecting JPEG quality factor values according to the input is another related research direction, In \cite{yang2021compression} Yang \textit{et al.} propose Highest Rank Selector (HRS) which for each original image, HRS will select the optimal compression versions as the input of the DNN model, which can significantly reduce the file size of input images while maintain the classification accuracy. After that, Xie \textit{et al.} developed AdaCompression \cite{li2019adacompress}, a JPEG quality factor values selector based on reinforcement learning agent. 

In the context of our research, a paper that closely aligns with our work was published in 2021 by Google Research \cite{50190}. The authors of this paper proposed a novel numerical method to achieve optimal quantization tables by using a combination of a third-order polynomial approximation of the quantization and a linear estimator to predict the compression ratio.  Their proposed algorithm demonstrated significant improvements in compression performance while maintaining high perceptual quality for human observers and preserving the performance of DNN-based models. However, this algorithm is limited to resized images with a fixed resolution, which may not be suitable for some real-world applications. Nonetheless, the success of their approach provides valuable insights into optimizing quantization tables for both human and machine perception.

\begin{figure*}[ht]
  \includegraphics[width=\textwidth]{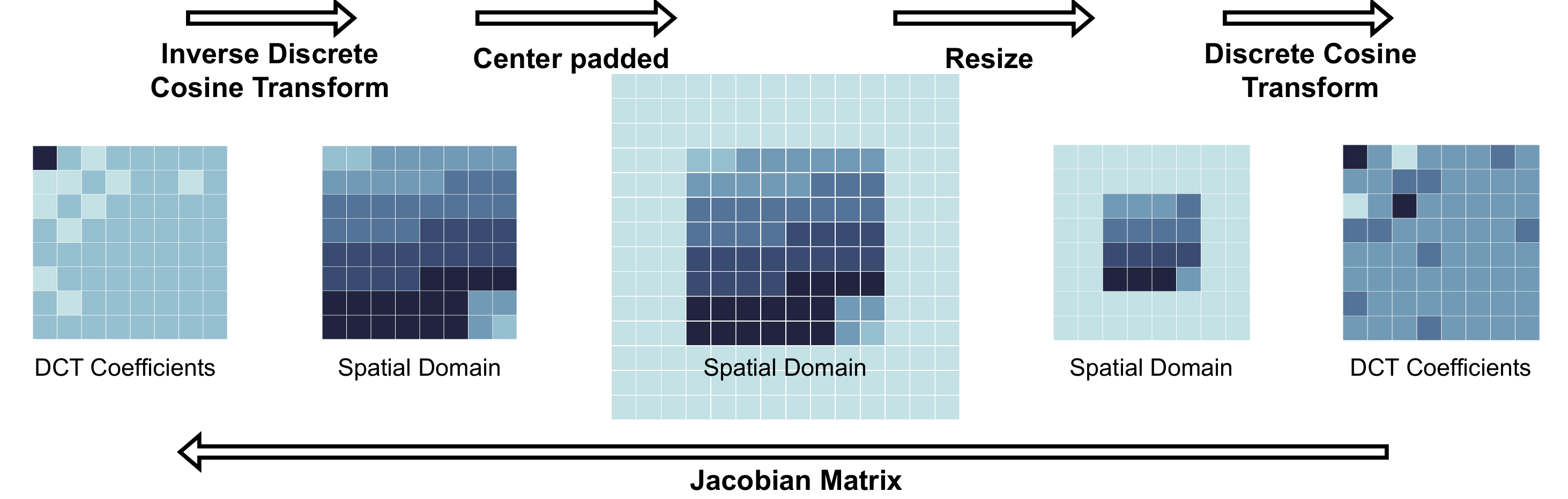}
  \caption{The example of how to get the Jacobian matrix J which will be used in Adaptive Sensitivity Mapping(ASM), Since all operations are linear, the resulting Jacobian matrix is independent to the input, which can easily get by Pytorch's autograd mechanism.}
  \label{fig:AdaptiveSenMapJacobian}
\end{figure*}

\section{PRELIMINARIES}
\label{sec:preliminaries}
JPEG is one of the most popular lossy image compression algorithms, whose mechanism is briefly explained in the following.

After converting the input RGB image $x$ with $m\times n$ spatial resolution into YCbCr color space, the JPEG encoder first pads each edge of $x$ into a multiplier of 8 and partition the padded image into $B$ non-overlapping $8\times8$ blocks. Each block is transformed into DCT domain by an $8\times8$ fast DCT transform, which yielding the DCT coefficients $x_f$. Then, $x_f$ will be uniformly quantized by an $8\times8$ quantization table $Q$, the resulting DCT indices $I$ are entropy coded by run-length coding and Huffman coding, which yields a sequence of run-length pairs $(r,s)$ followed by the non-zero amplitudes $a$, and a customized Huffman table $H$, respectively. According to the JPEG standard, Y channel is compressed by its own quantization table $Q_Y$ and Huffman table $H_Y$, while Cb and Cr channel share the same quantization table $Q_C$ and Huffman table $H_C$. 

The default image-independent JPEG pipeline is sub-optimal, because (i) the JPEG standard allows for custom quantization table and Huffman codewords, (ii) the compression results are largely determined by the quantization table. In order to reach better distortion compression rate trade-off, the following optimization problem can be formulated:

\begin{equation}
\label{eq:SDQFormula}
\min_{(r,s,a),H,Q} D[x,(r,s,a)_Q]\quad s.t\   R[(r,s),H]\leq R_T,
\end{equation}

where $D[x,(r,s,a)_Q]$ denotes the measure of human distortion between the original image $x$ and the reconstructed compressed image determined by $(r,s,a)$ and Q, which is evaluated using the peak signal-to-noise ratio (PSNR), $R[(r,s),H]$ denotes the compression rate obtained from $(r,s)$ and Huffman table $H$. In (\ref{eq:SDQFormula}), $R_T$ represents the rate budget.
The optimization problem mentioned above has been well studied in the last two decays. One of the state-of-the-art algorithms in the literature is the trellis-based iterative optimization algorithm credits to Yang \textit{et al.} \cite{YangSDQ}. Despite its superior trade-off, the results obtained by Yang \textit{et al.} only consider human distortion, which is not suitable for DNN accuracy. To tackle this issue, in this paper, we aim to develop a new image compression algorithm for both human and machine by formulating a multi-objective optimization problem for lossy image compression.

\section{THE PROPOSED METHOD}
\label{sec:ProbosedMethod}
\subsection{Towards Distortion Measure For Both Human and Machine}
\label{sec:problemformulation}
To take DNN accuracy into account, a new optimization objective will be introduced in this section. Without loss of generality, we only consider classification task in this paper. Given a DNN based image classifier $f(\cdot)\rightarrow \mathbb{R}^K$ with $K$  number of outputs, and a linear spatial interpolation operator $R(\cdot)$ that maps the any natural images with higher resolutions to an $m\times n$ resized image. The error rate $\mathbb{E}$ of $f$ for a dataset $X$ contains $N$ natural images $x_i$ with ground truth $GT_i$, $i\in[1,N]$, can be defined as:

\begin{align}
\mathbb{E} & =\frac{1}{N}\sum_{i=1}^N \mathbb{I}(f(R(x_i)), GT_i); \nonumber\\
  \mathbb{I}(\cdot) & =
    \begin{cases}
      0 & \text{if ${argmax}_i f(x_i)=GT_i$}\\
      1 & \text{if ${argmax}_i f(x_i)\neq GT_i$}\\
    \end{cases}   
    \ ~ i \in [1,N].
\end{align}

Since the error rate $\mathbb{E}$ defined above is non-differentiable, in practice, various differentiable surrogate loss functions $l$ are used so as to make the problem analytically solvable \cite{bartlett2006convexity}. In this work, we choose cross entropy loss as the surrogate loss to measure the distortion for DNN-based image classifiers, since it is widely adopted in DNN field. Then, our new optimization problem can be written as

\begin{equation}
\label{eq:MultiObjectOptProblem}
\min_{(r,s,a),H,Q} D[x,(r,s,a)_Q]; |\Delta L|\ \quad s.t\  R[(r,s),H] \leq R_T ,
\end{equation}

where $L\left(\cdot\right)$ is a composite function defined as $L\left(\cdot\right) = l(f(R(\cdot)))$, and $R_T$ denotes the target compression rate value. With the intention of unveiling the relations between distortion introduced by quantization over each DCT coefficients of compress images and the change of surrogate loss values.

First by Taylor expansion, the change of surrogate loss value can be upper bounded by:
\begin{align}
\label{eq:loss_increase}
|\Delta L| & = \left|L(x_f+\Delta x_f)-L(X_f)\right| = \left|\nabla L(x_f)\right|^T\Delta x_f+o(\|\Delta x_f\|) \nonumber\\
           & \leq \left|\nabla L(x_f)\right|^T\Delta x_f + \left|o(\|\Delta x_f\|)\right| \nonumber\\
           & =\left|\sum_{i=1}^{64}\frac{d L}{d C_i^T}\Delta {x_f}_i\right|+\left|o(\|\Delta x_f\|)\right| \nonumber\\
           & \leq \sum_{i=1}^{64} \left|\frac{d L}{d {x^T_f}_i}\Delta{x_f}_i\right| + \left|o(\|\Delta x_f\|)\right|.
\end{align}

Using Cauchy-Schwarz inequality and squaring both sides of formula \ref{eq:loss_increase} yields:

\begin{align}
\label{eq:loss_upperbound}
\Delta L^2 & \leq \left(\sum_{i=1}^{64}\sqrt{\sum_{j=1}^B\left(\frac{\partial L}{\partial {x_f}_{i,j}}\right)^2\cdot\sum_{j=1}^B\Delta {x_f}_{i,j}^2}\right)^2+\left|o\left(\|\Delta x_f\|^2\right)\right| \nonumber\\
           & \leq 64\sum_{i=1}^{64}\sum_{j=1}^B\left(\frac{\partial L}{\partial {x_f}_{i,j}}\right)^2\cdot\sum_{j=1}^B\Delta {x^2_f}_{i,j} + o\left(\|\Delta x_f\|^2\right).
\end{align}   

Then, the square of change of surrogate loss caused by distortion on $i^{th}$ coefficient $\Delta {x_f}_i$ can be written as:

\begin{align} 
&\Delta L^2_i  \leq \sum_{j=1}^B\left(\frac{\partial L}{\partial {x_f}_{i,j}}\right)^2\cdot\|\Delta {x_f}_i\|^2+o\left(\|\Delta {x_f}_i\|^2\right) \\
&\frac{\Delta L^2}{\|\Delta {x_f}_i\|^2}  \leq \sum_{j=1}^B\left(\frac{\partial L}{\partial {x_f}_{i,j}}\right)^2+o(1),
\end{align} 
where $\|\Delta {x_f}_i\|^2 = \sum_{j=1}^B \Delta {x^2_f}_{i,j}$. We now define $S_i=\sum_{j=1}^B\left(\frac{\partial L}{\partial {x_f}_{i,j}}\right)^2$ as the sensitivity.

Now, the square of the change in surrogate loss value resulting from distortion in DCT coefficients can be upper bounded by $\sum_{i=1}^M S_i\cdot \|\Delta {x_f}_i\|^2$. However, the sensitivity depends on the specific DNN model, input image, and ground truth, which may not always be accessible or affordable for a image compression encoder, therefore the sensitivity has to be estimated offline.

\subsection{Offline Estimation Of The Sensitivity And Adaptive Sensitivity Mapping}\label{sec:OfflineEstimationAdaptiveSenMap} 
In order to overcome this challenge raised in the previous section, we estimate the sensitivity offline by randomly selecting $N'$ images with the same resolution from the training set. Then the sensitivity can be estimated by the sample mean of the sensitivity over the $N'$ images.
\begin{equation}
\hat{S_i}=\frac{1}{N'}\sum_{n=1}^{N'}\sum_{j=1}^B\left(\frac{\partial L}{\partial {x^n_f}_{i,j}}\right)^2,\ i\in[1,64].
\end{equation}

Notice that $L$ is a composite function of the surrogate loss $l$, DNN $f$, and liner interpolation operator $R$. To match the resolution of the DNN input, different $R$ will be applied to scale the images of various resolutions, which implies that the estimated sensitivity $\hat{S}$ varies for the images with different resolutions. Since natural images has infinite different resolutions, estimate the sensitivity from a sample of real images for all resolutions is not feasible.

We fill this gap by introducing adaptive sensitivity mapping(ASM). ASM is designed to approximate the sensitivity of images with a particular resolution using the estimated sensitivity of images at a lower resolution, which relies on two moderate assumptions: 
\begin{enumerate}
    \item[] Assumption (i): The gradient on each pixel are i.i.d from a certain distribution.
    \item[] Assumption (ii): The random variables of ${G_f}_i=\frac{\partial L}{\partial {X^n_f}_{i}}, \ i\in[1,64]$ are zero-mean.
\end{enumerate}

As Assumption (i) is widely accepted in literature \cite{glorot2010understanding, he2015delving, ye2024bayes, 9817678}, we only verify the Assumption (ii) in the appendix. Since the DCT transform matrix is orthonormal, based on the Assumption (i), ${{G_f}_i\ i\in[1,64]}$ are independent to each other.

 Without loss of generality, we assume an original image $x$ with spatial resolution $m\times m$, and the DNN-based model take resized images with fixed resolution $n\times n$ as input. Then an $8\times8$ block in the resized image corresponds to a $\lfloor\frac{8m}{n}\times\frac{8m}{n}\rceil$  block in the original image. To estimate the sensitivity on the original image $\hat{S_i}'$, we first apply a series of linear operations on an $8\times8$ DCT coefficients. Firstly, we transform an $8\times8$ DCT coefficients to $8\times8$ spatial block, then center-pad it to $\lfloor\frac{8m}{n}\times\frac{8m}{n}\rceil$. Next, we resize the padded block back to $8\times8$  and transform it back to the DCT domain. Since, all these operations are linear, making its Jacobian matrix $J$ is independent of the input. Figure \ref{fig:AdaptiveSenMapJacobian} illustrates the process of obtaining the Jacobian matrix. Thanks to PyTorch \cite{PyTorch}’s autograd mechanism, The Jacobian matrix $J$ is easy to obtain.

\begin{figure} 
    \centering
  \subfloat[Estimated sensitivity of image with small side equal to 512.]{%
       \includegraphics[width=1\linewidth]{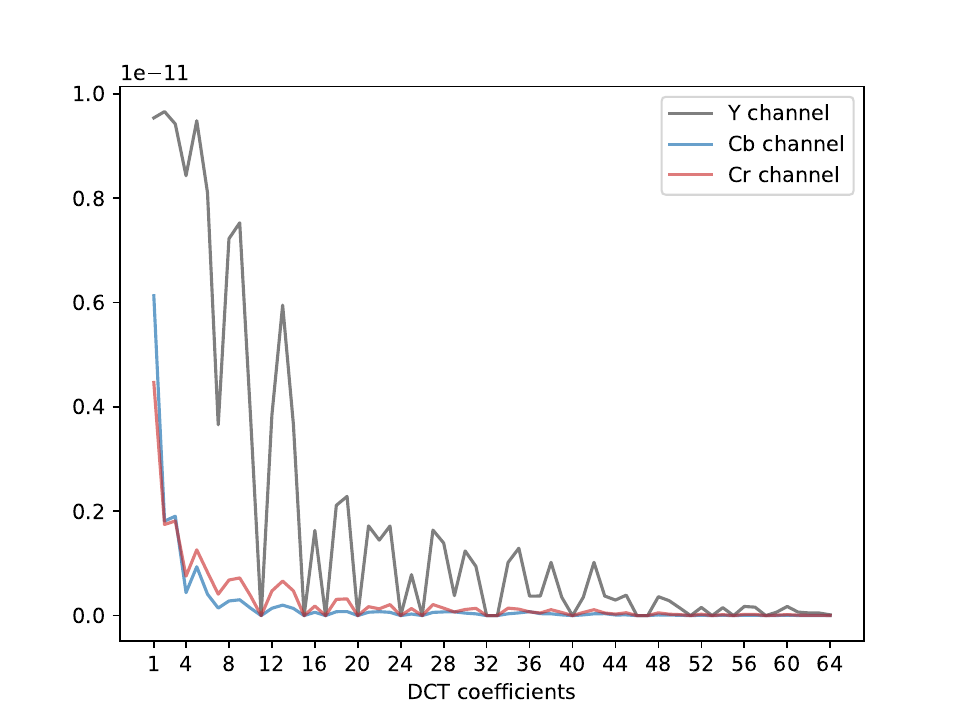}
       \label{fig:AdaptSenMapCompare_a}}
    \hfill
  \subfloat[Approximated estimated sensitivity of image with small side equal to 512 using ASM to map from estimated sensitivity of images with resolution $224\times224$.]{%
        \includegraphics[width=1\linewidth]{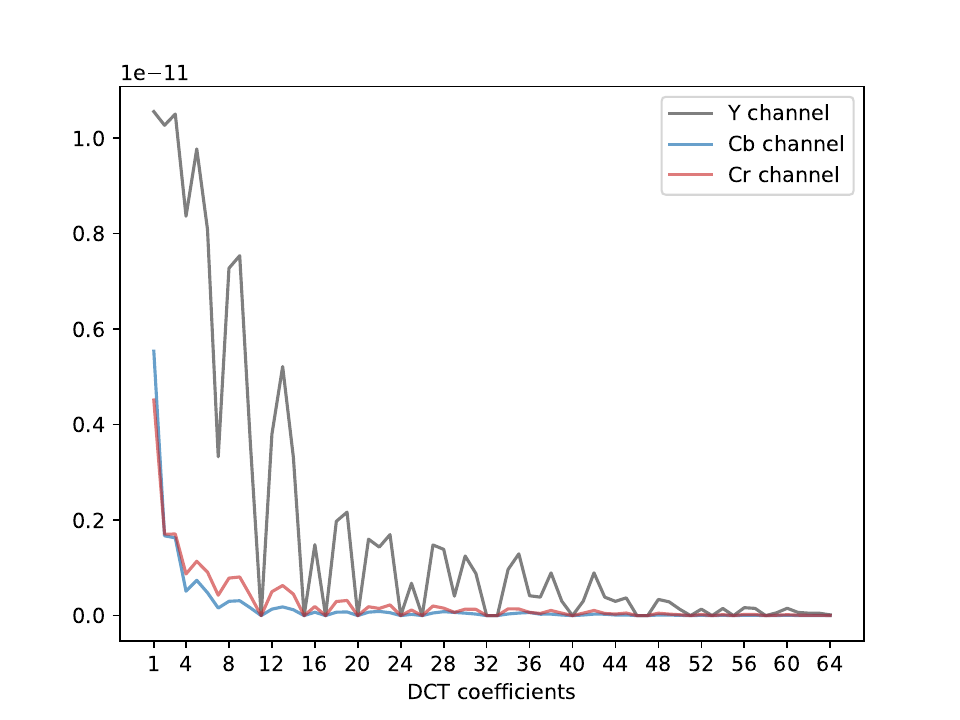}
        \label{fig:AdaptSenMapCompare_b}}
    \\
  \caption{Fig \ref{fig:AdaptSenMapCompare_a} and Fig \ref{fig:AdaptSenMapCompare_b} shows that the ASM can accurately approximate the estimate sensitivity of image with higher resolution.}
    \label{fig:AdaptSenMapCompare}
\end{figure}

Then the approximated random variable of gradient on the DCT coefficient of original image ${G_f}^o_i\ i\in[1,64]$ can be calculated as:
\begin{equation}
{G_f}^o_i=\sum_{k=1}^{64}J_{i,k}{G_f}_{k},\ i\in[1,64].
\end{equation}

Based on assumption $(2)$, we can rewrite the estimated sensitivity as
\begin{equation}
\hat{S_i}' = \frac{{N'}B-1}{{N'}B}var({G_f}_i);\ i\in[1,64].
\end{equation}

Then the estimated sensitivity on original image can be written as
\begin{align}
\hat{S_i}' &= \frac{{N'}B-1}{{N'}B}var(\sum_{k=1}^{64}J_{i,k}{G_f}_k) = \frac{{N'}B-1}{{N'}B}\sum_{k=1}^{64}J_{i,k}^2 var({G_f}_k) \nonumber\\ 
&=\sum_{k=1}^{64}J_{i,k}^2 \hat{S_i}\ i\in[1,64].
\end{align}

Figs \ref{fig:AdaptSenMapCompare}  illustrate the estimated sensitivity $\hat{S}$ for AlexNet on images with shorter side equal to 512 from the ImageNet Training set, as well as the approximated estimated sensitivity obtained using ASM to map from the resized images, which were first scaled to a resolution with shorter side equal to 256, then center-cropped to $224\times 224$. Our analysis indicates that ASM can produce a reasonably accurate approximation of sensitivity.

\subsection{Human and Machine-Oriented Error}
From equation \ref{eq:loss_upperbound} and sensitivity $\hat{S}$ estimated in section \ref{sec:OfflineEstimationAdaptiveSenMap}, we can upper bound the square of the loss change caused by quantization as:
\begin{equation}
\label{eq:SWE}
 \Delta L^2_i  \leq D^m({x_f}_i,q)= \sum_{k=1}^{64}\hat{S}_i({x_f}_i-I_iQ_i)^2.
\end{equation}

By replacing the optimization objective of surrogate loss in formula \ref{eq:MultiObjectOptProblem} with (\ref{eq:SWE}) we can rewrite the multi-objective optimization problem as:

\begin{equation}
\label{eq:Dmoo}
\min_{(r,s,a),H,Q}D[x,(r,s,a)_Q];\ D^m({x_f}_i,Q) \quad s.t\  R[(r,s),H] \leq R_T,
\end{equation}

where $D^m({x_f}_i,Q)$ is defined as machine distortion. By defining the associated Lagrangian function for the objectives as:

\begin{equation}
\label{eq:OptimzationHMOE}
\min_{(r,s,a),H,Q} \sum_{i=1}^{64}(1+\lambda\cdot\hat{S}_i)({x_f}_i - I_iQ_i)^2,\ s.t\ R[(r,s),H]\leq R_T,
\end{equation}

where $R_T$ is the compression rate budget for the compressed images. Now, We define Human and Machine-Oriented Error (HMOE), a new distortion measure for both human and machine, as:
 \begin{equation}
 {HMOE} = \sum_{i=1}^{64}(1+\lambda\cdot\hat{S}_i)({x_f}_i - I_iQ_i)^2,
 \end{equation}
where the Lagrangian multiplier $\lambda$ is a predefined parameter that represents the trade-off of human distortion for machine distortion.  Notice that by minimizing ${HMOE}$ subject to a rate constraint will in turn reduce the squared surrogate loss change and MSE.

Since the HMOE is independent to the compression algorithm, the problem formulated in subsection \ref{sec:problemformulation} can be solved by replacing the MSE with the HMOE in any off-the-shelf image compression algorithm for HVS. At this point, we have all components for developing an image compression algorithm for both human and machine, which will be represented in the next section.

 \subsection{Human And Machine Oriented Soft Decision Quantization (HMOSDQ)}
With the help of the HMOE introduced in the previous section, in this paper, we choose the Soft Decision Quantization (SDQ) \cite{YangSDQ} as it can achieve superior trade-off between rate and distortion by iteratively optimizing the run-length coding, Huffman coding and quantization table. Since the Huffman table is determined by the probability distribution of run-length pairs $P^{r,s}$, the Huffman table H will be replaced with $P^{r,s}$ in the optimization process. Now, we briefly discuss the SDQ algorithm as follows:
\begin{enumerate}
    \item We estimate $P^{r,s}_0$ from an input image $x$ and initial quantization table $Q_0$.
    \item\label{item:SDQAlgOptStep1}Fix $Q_t$ and $P^{r,s}_t$, for any $t\geq0$, for each DCT block we find an optimal sequence of $\left(r,s,a\right)$ that achieve the following minimum:
    \begin{equation}
    \min_{r,s,a} D\left[x_0, \left(r,s,a\right)_{Q_t}\right]+\beta R\left[\left(r,s\right)_{Q_t}, P^{r,s}_t\right],
    \end{equation}
    by trills quantization.
    \item\label{item:SDQAlgOptStep2}Fix $\left(r_t,s_t,a_t\right)$, update $Q_t$ and $P^{r,s}_t$, in order to achieve the following minimum:
    \begin{equation}
    \min_{Q,P^{r,s}} D\left[x_0, \left(r_t,s_t,a_t\right)_Q\right]+\beta R\left[\left(r_t,s_t\right)_Q, P^{r,s}\right].
    \end{equation}
    \item Repeat step \ref{item:SDQAlgOptStep1} and step \ref{item:SDQAlgOptStep2} until convergence or reach stop criteria.
\end{enumerate}

As the original SDQ is designed for monochrome image, we extend it for colour images in order to be compatible with DNNs' input. Since Y channel has its own quantization table and Huffman table, as mentioned in section \ref{sec:preliminaries}, original SDQ can be directly applied after replacing MSE with HMOE. While for the Cb and Cr channels, besides replacing MSE with HMOE, the algorithm has to optimize both chroma channels simultaneously, which is described as follows:
\begin{enumerate}
    \item We estimate ${P^{r,s}_c}_0$ from both Cb channel $x^{b}$ and Cr channel  $x^{r}$  of an input image $x$ and initial quantization table ${Q^c}_0$ for both $x^{b}$ and $x^{r}$.
    \item\label{item:SDQAlgOptStep1_c}Fix ${Q^c}_t$ and ${P^{r,s}_c}_t$, for any $t\geq0$, for each DCT block we find optimal $\left(r^r,s^r,a^r\right)$ and $\left(r^b,s^b,a^b\right)$ for $x^{b}$ and $x^{r}$ respectively to achieve the following minimum:
    \begin{align}
    \min_{\substack{r^b,s^b,a^b,\\r^r,s^r,a^r}}\ 
    &HOME_b\left[{x^b_0}, \left(r^b,s^b,a^b\right)_{Q^c_t}\right]+\beta R\left[\left(r^b,s^b\right)_{Q^c_t}, {P^{r,s}_c}_t\right] \nonumber\\+
    &HOME_r\left[{x^r_0}, \left(r^r,s^r,a^r\right)_{Q^c_t}\right]+\beta R\left[\left(r^r,s^r\right)_{Q^c_t}, {P^{r,s}_c}_t\right],
    \end{align}
    by trills quantization.
    \item\label{item:SDQAlgOptStep2_c}Fix $\left(r^b_t,s^b_t,a^b_t\right)$ and $\left(r^r_t,s^r_t,a^r_t\right)$, update $Q^c_t$ and ${P^{r^r,s^r}_c}_t$, in order to achieve the following minimum:
    \begin{align}
    \min_{Q^c,P_c^{r,s}}
    &HOME_b\left[x^b_0, \left(r^b_t,s^b_t,a^b_t\right)_{Q^c}\right]+\beta R\left[\left(r^b_t,s^b_t\right)_{Q^c}, P_c^{r,s}\right]\nonumber\\+
    &HOME_r\left[x^r_0, \left(r^r_t,s^r_t,a^r_t\right)_{Q^c}\right]+\beta R\left[\left(r^b_t,s^b_t\right)_{Q^c}, P_c^{r,s}\right].
    \end{align}
    \item Repeat step \ref{item:SDQAlgOptStep1_c} and step \ref{item:SDQAlgOptStep2_c} until convergence or reach stop criteria.
\end{enumerate}
Now we get a JPEG-compliant image compression algorithm for both human and machine. In the next section, we evaluate the performance of proposed method though as set of experiments.

\section{EXPERIMENTS}
\label{sec:experiments}
This section starts with the experimental setup. After that, we present the results of the rate-distortion-accuracy performance on two subsets of ImageNet validation set with two DNN models. Finally, the ablation study proofs the effectiveness of HMOSDQ we proposed in the paper come form.
\subsection{Experimental Setup}
Two subsets of the ImageNet validation set will be chosen: one with its images whose shorter side are in the range 496 to 512, and the other one with those in the range 376 to 384, denoted as $\left(\mathcal{S}_{512}\right)$ and $\left(\mathcal{S}_{384}\right)$, respectively. We use OPs\cite{Ops} to initial quantization table for each image, then it will be compressed by HMOSDQ which yields the BPP and the PSNR. Next, the smaller edge of images will be padded to 512 or 384 according to the subset it belongs to. Then, we follow the standard prepossessing of Imagenet: Scaling the image such that the smaller edge is equal to 256, then center crop them to $224\times224$. Finally, the preprocessed images are inferenced by VGG-16 and Alexnet to get the accuracy, which is defined as accuracy=$1-\mathbb{E}$.

In order to establish various trade-offs between distortion and accuracy, three different $\lambda$ $\left(10^{11},\ 10^{12},\ 10^{13}\right)$ are chosen. 

The low-resolution sensitivity $\hat{S}$ is estimated based on 10,000 randomly sampled resized images with resolution $224\times224$ from the ImageNet training set. Then the estimated sensitivity of high-resolution images in the two subsets we selected are yielded from $\hat{S}$ via ASM.

The compression rate (measured in BPP), distortion (measured in PSNR), and accuracy are averaged over all images in the selected subset and reported in the following subsection.

\subsection{Results}

\begin{figure}
  \centering  
\includegraphics[width=\columnwidth]{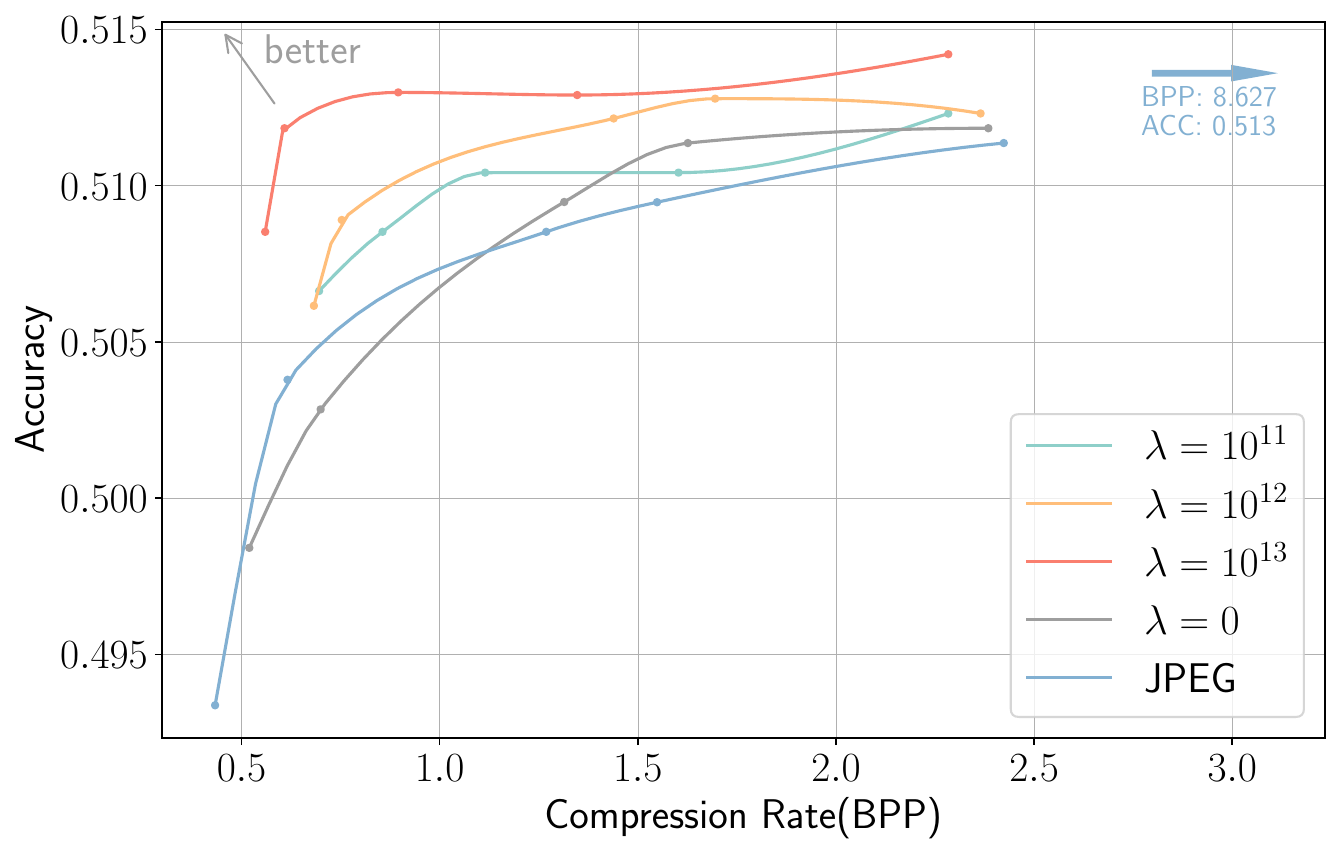}
  \centering  
\includegraphics[width=\columnwidth]{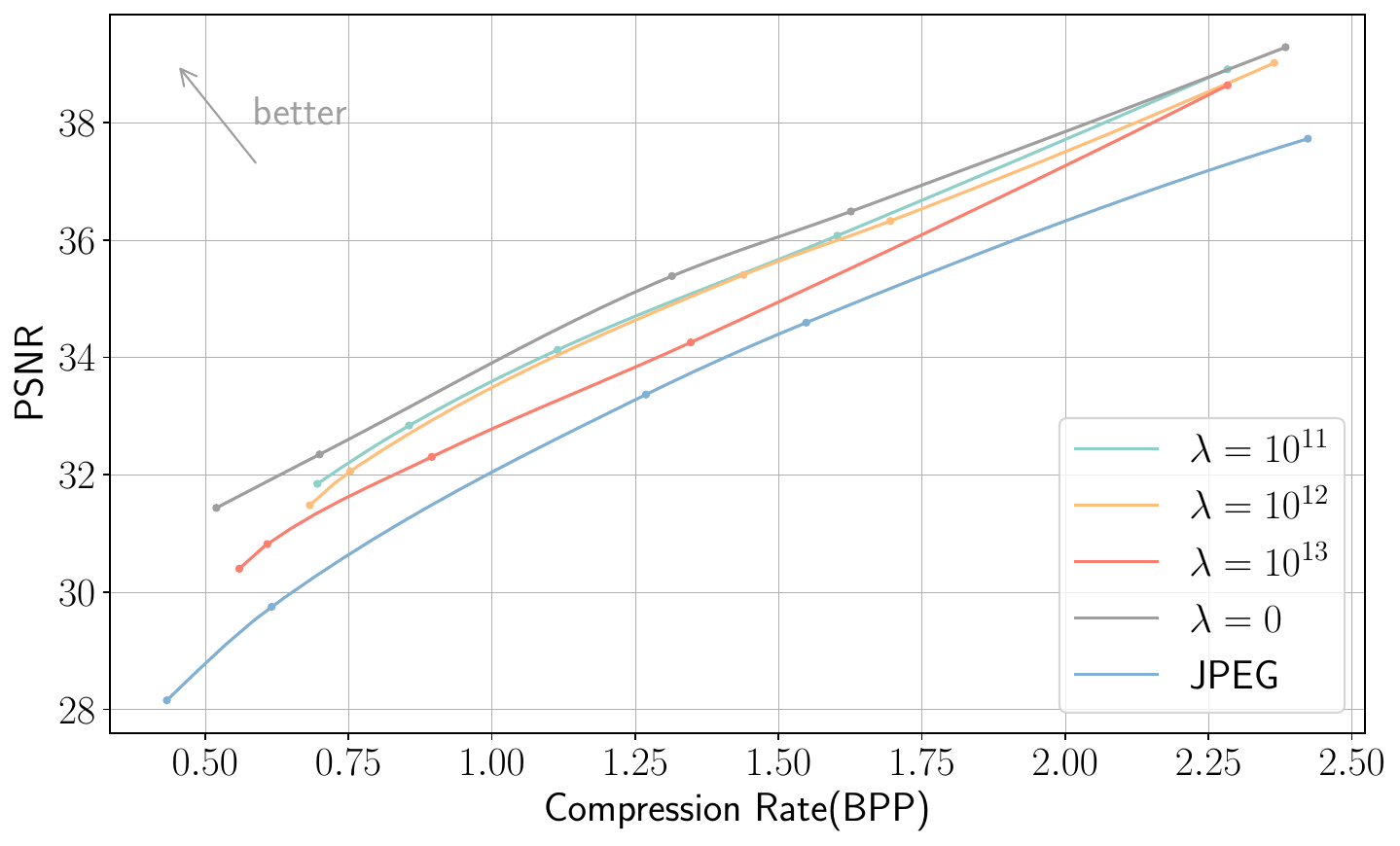}
  \caption{Average PSNR and validation accuracy of Alexnet on subset of images with smaller edge within the range 496 to 512.}
  \label{fig:Alexnet512Results}
\end{figure}

 \begin{figure}
  \centering  
  \includegraphics[width=\columnwidth]{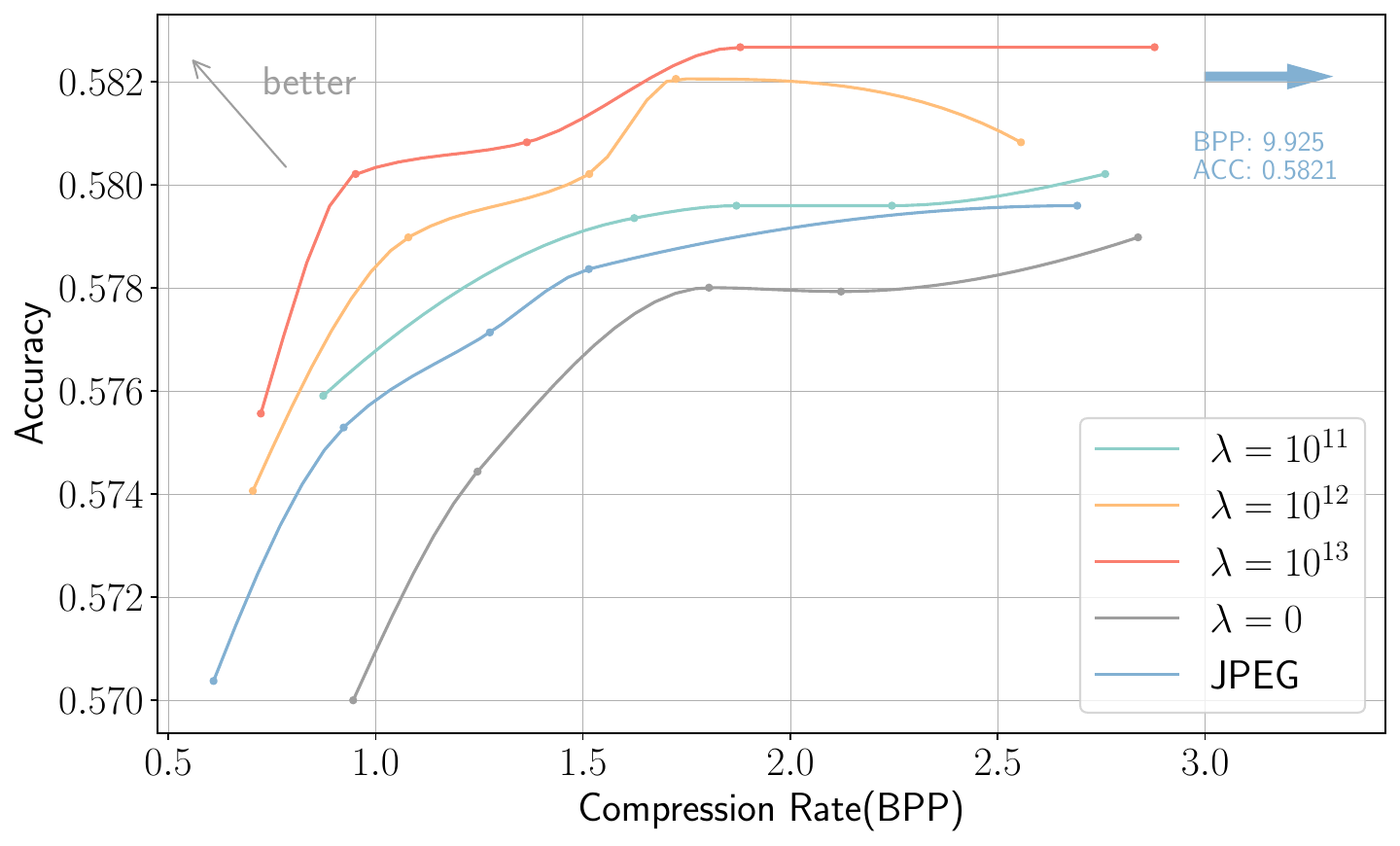}
  \centering  
  \includegraphics[width=\columnwidth]{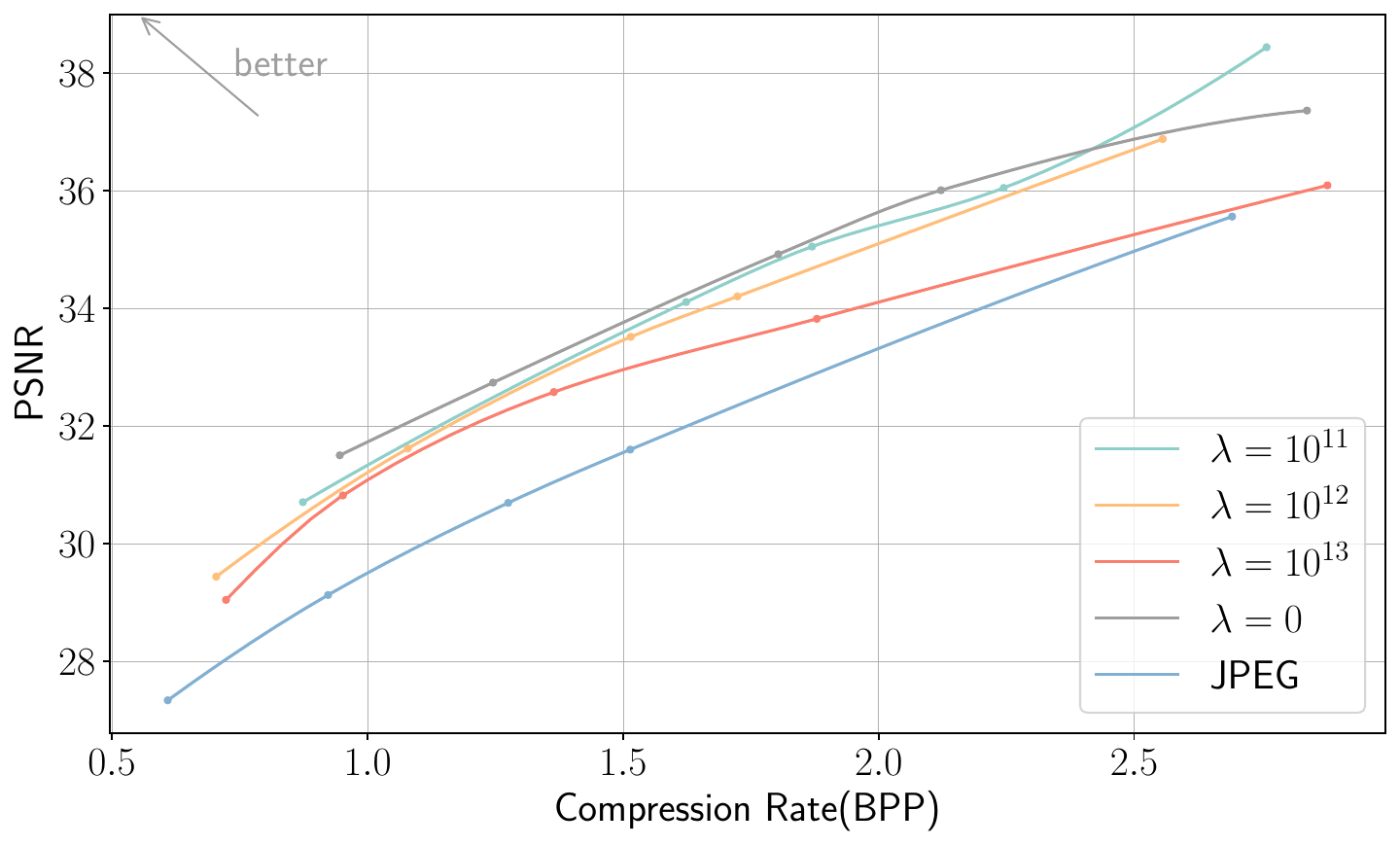}
  \caption{Average PSNR and validation accuracy of Alexnet on subset of images with smaller edge within the range 376 to 384.}
  \label{fig:Alexnet384Results}
\end{figure}

\begin{figure}
  \centering  
  \includegraphics[width=\columnwidth]{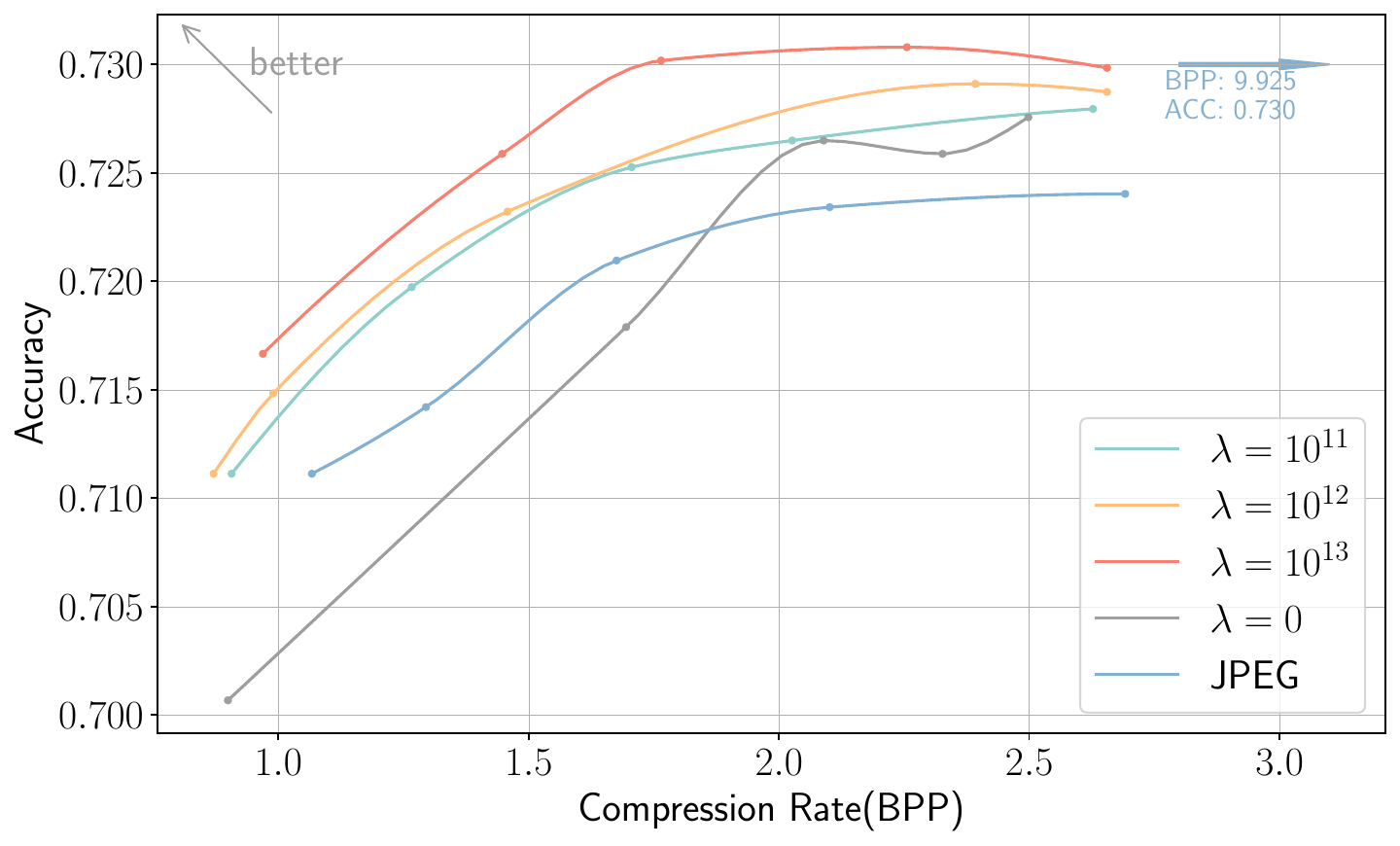}
  \centering  
  \includegraphics[width=\columnwidth]{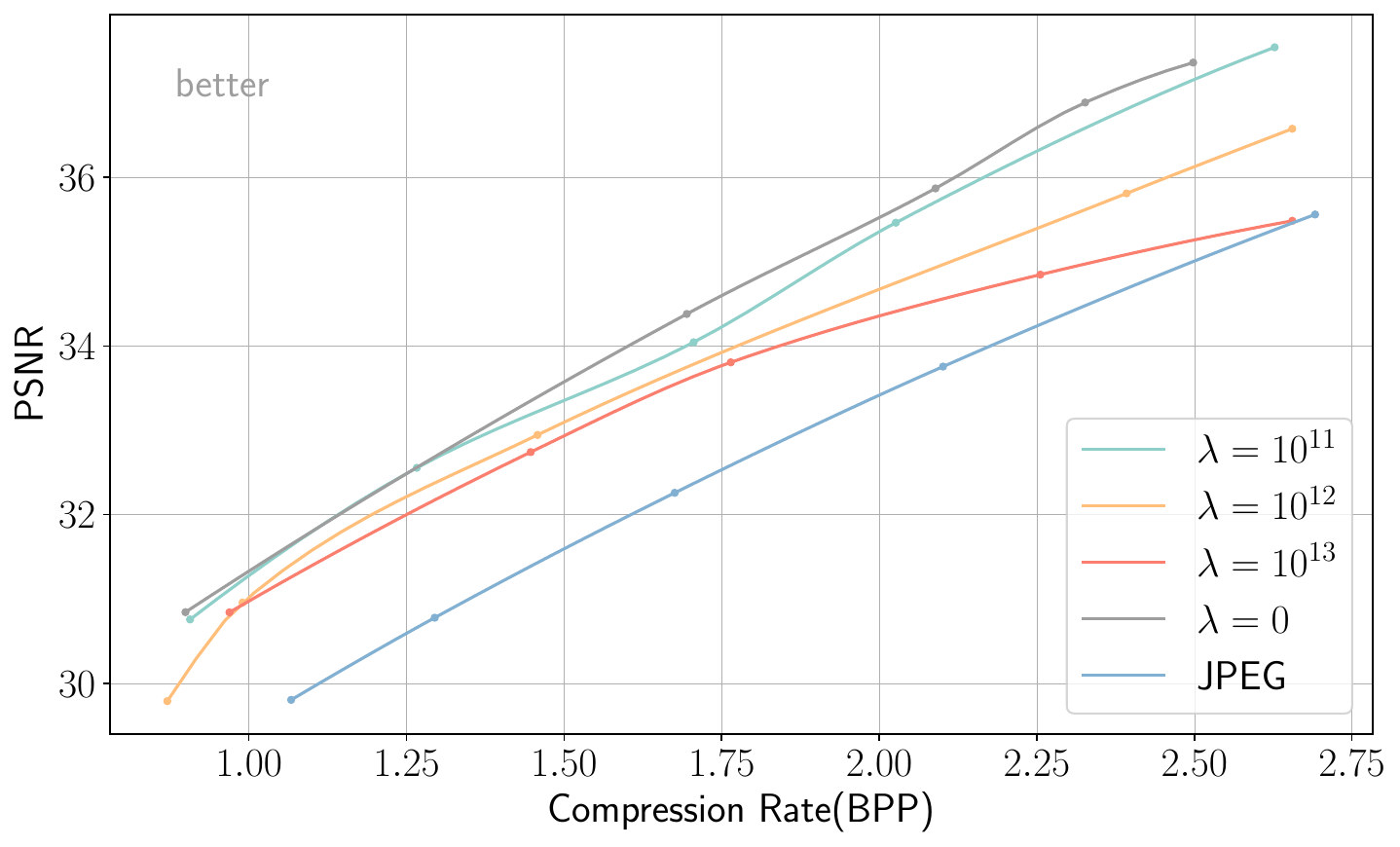}
  \caption{Average PSNR and validation accuracy of VGG-16 on subset of images with smaller edge in the range 376 to 384.}
  \label{fig:VGG16384Results}
\end{figure}

 \begin{figure}
  \centering  
  \includegraphics[width=\columnwidth]{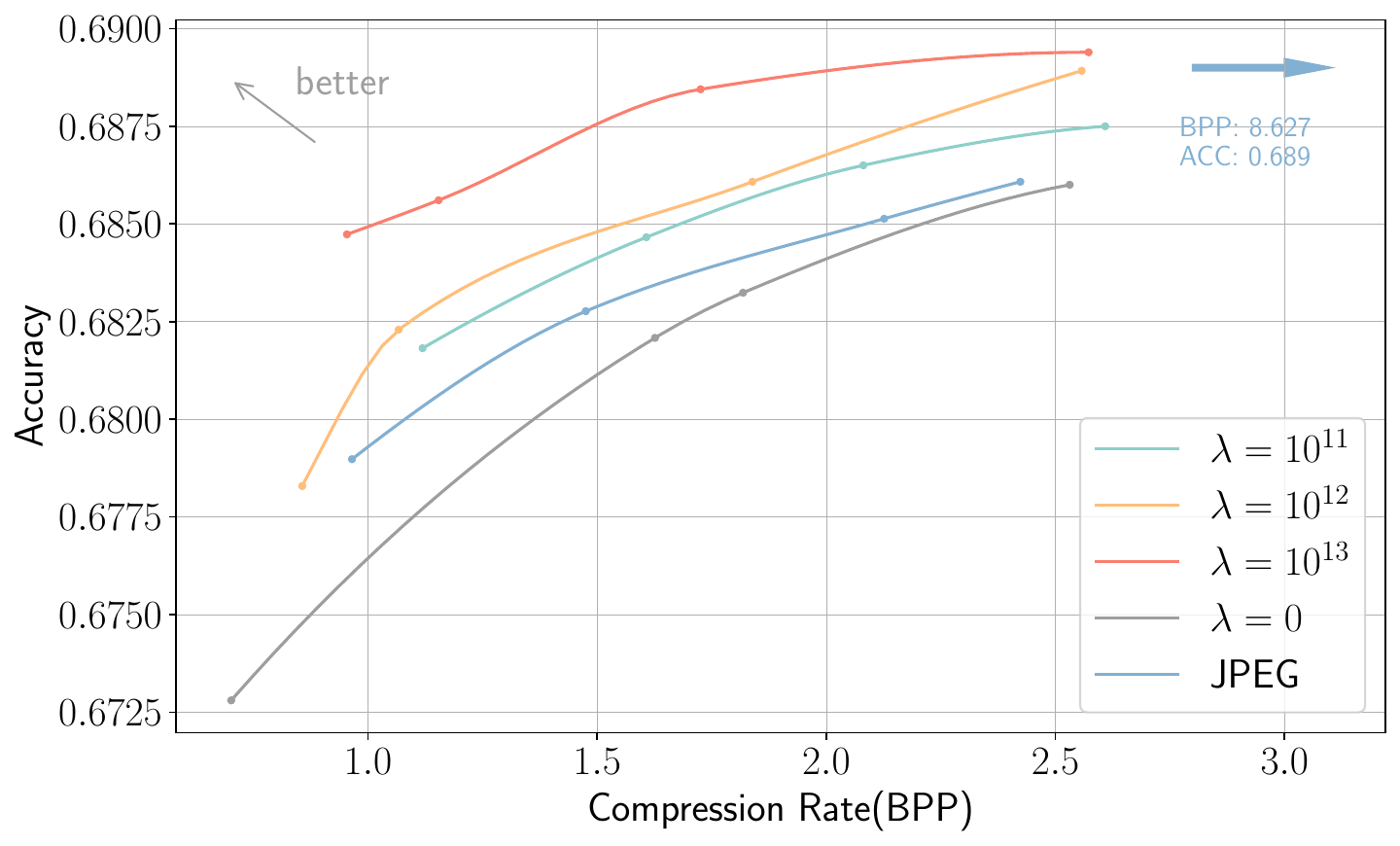}
  \centering  
  \includegraphics[width=\columnwidth]{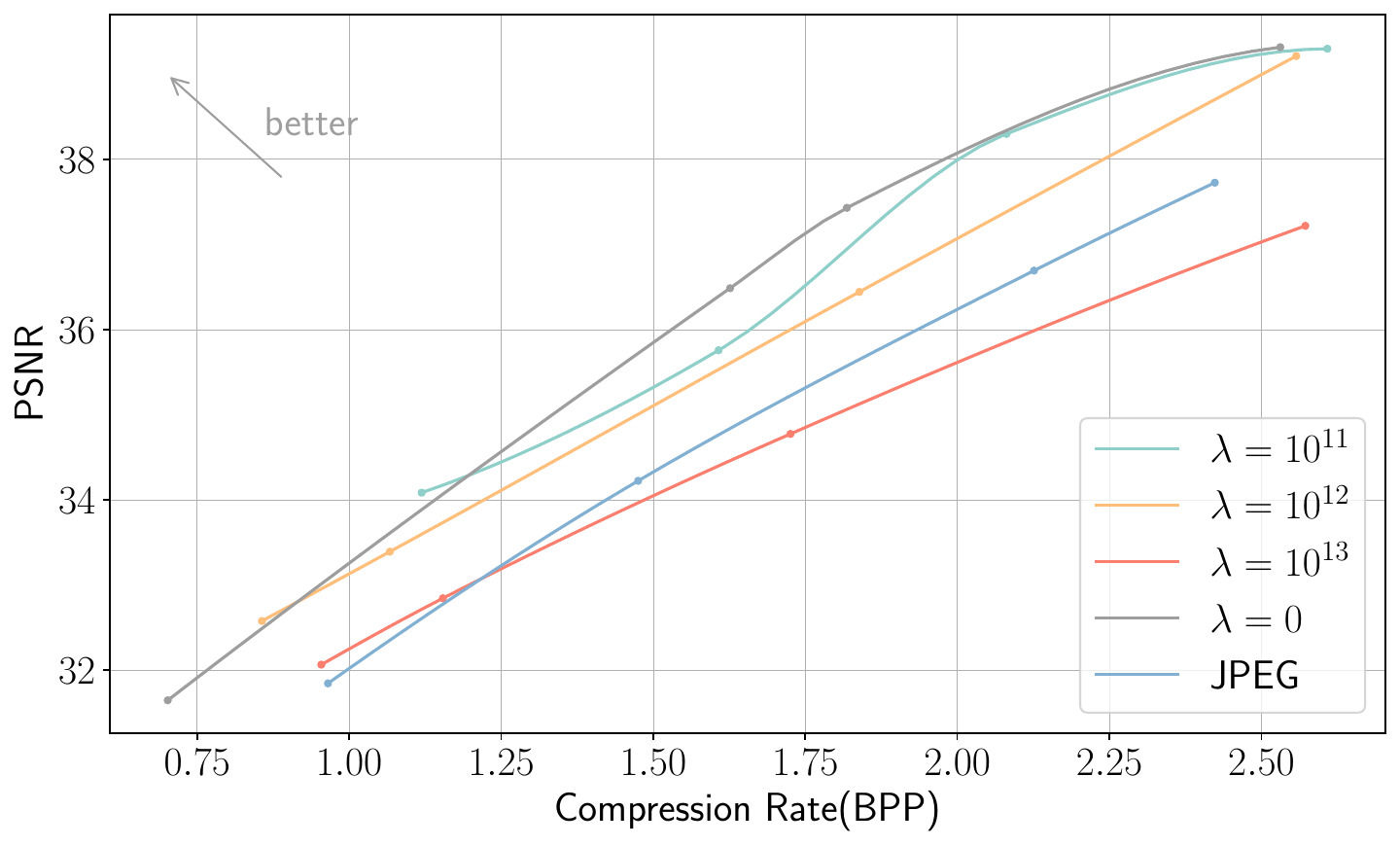}
  \caption{Average PSNR and validation accuracy of VGG-16 on subset of images with smaller edge within the range 496 to 512.}
  \label{fig:VGG16512Results}
\end{figure}
In this section, we analyze the experimental results of our proposed algorithm in terms of rate-accuracy and rate-distortion performance separately. All the results are presented graphically in figures (\ref{fig:Alexnet512Results}, \ref{fig:Alexnet384Results}, \ref{fig:VGG16384Results}, \ref{fig:VGG16512Results}).
  
 We start with evaluating the rate-distortion performance of our proposed algorithm. The results show that, for three trade-offs with different $\lambda$s established in the previous section,  we can always achieve a higher PSNR within the same compression rate, except for the results of VGG-16 with $\lambda=10^{13}$ evaluate on the images in $\mathcal{S}_{512}$, as showed figure \ref{fig:Alexnet512Results}. Specifically, when we use HMOSDQ with sensitivity of VGG-16 on images from $\mathcal{S}_{384}$, we can obtain 2.1 dB gain in PSNR within same compression rate of 2.69 BPP. 

 Next, we turn our attention to the rate-accuracy trade-off.  Our results show that HMOSDQ consistently provides a better trade-off compared to the JPEG baseline. As shown in figure \ref{fig:Alexnet512Results}, we can reduce the compression rate by more than 9.6 times with no compromise in accuracy, or equivalently improve the validation accuracy by  $0.81\%$ at 0.61 BPP. 
 
 Last but not least, our experimental results confirm that with different $\lambda$, different trade-offs between accuracy and distortion can be established. Specifically, larger value of $\lambda$ will tilt the trade-off towards accuracy. 

Since the original SDQ algorithm can already achieve a superb distortion-rate trade-off, it is important to understand whether the accuracy gain achieved with HMOSDQ is solely due to the new distortion measure HMOE proposed in our paper. In the next section, we conduct an ablation study to investigate the effectiveness of HMOE.

\subsection{Ablation Study}
In order to  answer the question raised at the end of the previous section, an ablation study is conducted to evaluate the trade-off between the accuracy and compression rate of Alexnet and VGG-16 on the images compressed by SDQ by setting $\lambda=0$. The results are presented in figures (\ref{fig:Alexnet512Results}, \ref{fig:Alexnet384Results}, \ref{fig:VGG16384Results}, \ref{fig:VGG16512Results}). The original SDQ algorithm could not match the accuracy-rate trade-off achieved by HMOSDQ. In fact, in most cases, the original SDQ performed even worse than the JPEG compression algorithm in accuracy. These results confirm that the accuracy gain achieved with HMOSDQ is indeed due to the new distortion measure HMOE proposed in our paper.

\section{CONCLUSION AND FUTURE WORKS}
\label{sec:ConclusionAndFutureWorks}
In this paper, in order to improve the DNN-based image classifier performance against lossy image compression, we first derive the sensitivity, which harness the upper bound of the squared change of the surrogate loss caused by distortion applied to DCT coefficients in quantization stage. To overcome the sensitivity's dependence on input image, image resolution and DNN models, we proposed adaptive sensitivity mapping and estimated the sensitivity offline using sample mean. We then introduce a novel distortion measure, called Human and Machine Oriented Error (HMOE), which takes both human distortion and machine distortion into account. After that, by extending the SDQ to accommodate color images, and replacing the Mean Square Error (MSE) in original Soft Decision Quantization (SDQ) with HMOE, we get a lossy image classification algorithm for both human and machine dubbed HMOSDQ. Finally Our experimental results and ablation study demonstrate the effectiveness of the proposed distortion measure in improving the accuracy of the DNN-based image classifier under lossy image compression.

In the future, we plan to investigate the performance of HMOSDQ on more computer vision tasks like object localization or semantic segmentation. Another promising direction would be to extend the HMOE for other lossy coding formats of different contents like MP3, and H.264 in audio, and video contents, respectively.


\ifCLASSOPTIONcaptionsoff
  \newpage
\fi

\bibliography{main}
\bibliographystyle{IEEEtran}

%







\appendix

\section{Verify the distribution of gradient of DCT coefficients are zero-mean}

\begin{figure}[h!]
  \centering  
  \includegraphics[width=\columnwidth]{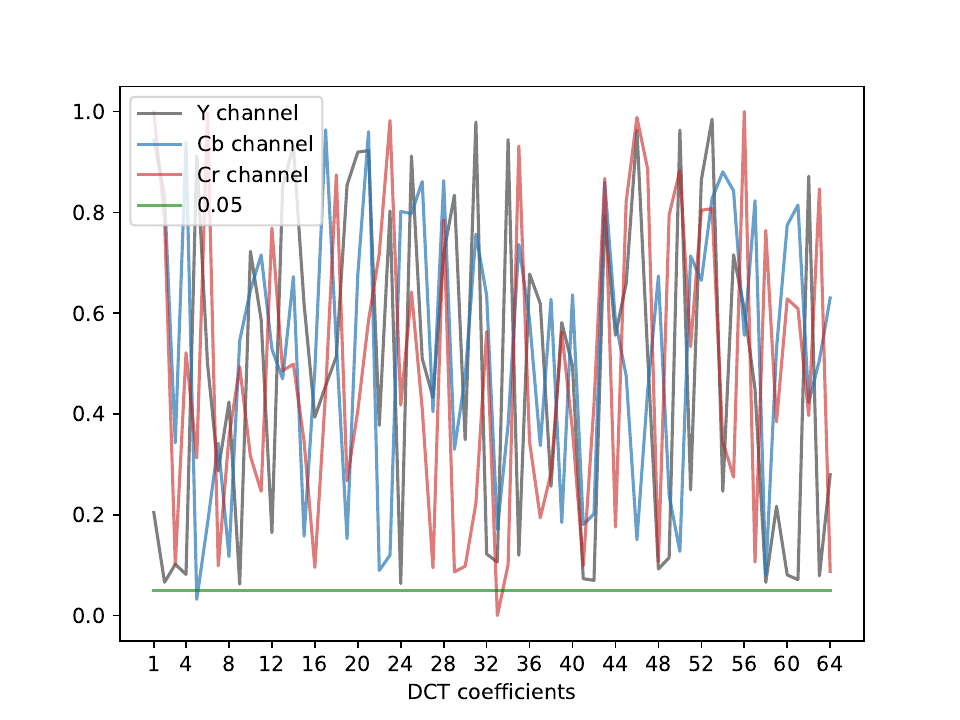}
  \caption{p-value in the t-test}
  \label{fig:Pval}
\end{figure}
T-test will be used to to verify the Assumption (ii): The random variables of ${G_f}_i=\frac{\partial L}{\partial {X^n_f}_{i}}, \ i\in[1,64]$ are zero-mean. in section \ref{sec:OfflineEstimationAdaptiveSenMap}.
Formally, we set up the following hypothesis testing:
\begin{enumerate}
    \item Null hypothesis $H_0$: the gradient of $i^{th}$ DCT coefficient ${G_f}_i$ of $8\times8$ DCT block follow from a zero-mean distribution.
    \item Alternative hypothesis $H_A$: the gradient of $i^{th}$ DCT coefficient ${G_f}_i$ of $8\times8$ DCT block follow from a non-zero-mean distribution.
\end{enumerate}
We apply T-test with significance level $\alpha=0.05$, figure \ref{fig:Pval} shows the respective p-value for the gradient of each coefficient $i\in[1,64]$. The Null hypothesis will be rejected if p-value is less than $\alpha=0.05$, and accept otherwise. Only the Null hypothesis of ${G_f}_{33}$ in Cr channel and  ${G_f}_5$ in Cb channel are rejected, thus based on T-test, we can conclude that ${G_f}_i\ i\in[1,64]$ follow zero-mean distribution.

\end{document}